\documentclass{article}
\usepackage{ijcai20}
\usepackage{algorithm}
\usepackage{algorithmic}
\usepackage{amsmath}
\usepackage{comment}
\usepackage{times}

\usepackage{soul}
\usepackage{url}
\usepackage[hidelinks]{hyperref}
\usepackage[utf8]{inputenc}
\usepackage[small]{caption}
\usepackage{graphicx}
\usepackage{amsmath}
\usepackage{booktabs}
\usepackage{xspace}

\newcommand{\algo}{\texttt{GUISE}\xspace}
\usepackage{tikz}
\newif\ifcomments
\commentstrue
\ifcomments
\newcommand{\authorcomment}[2]{\tikz[baseline=(X.base)]\node [draw=#1,fill=#1!40,semithick,rectangle,inner sep=2pt, rounded corners=3pt] (X) {#2};}

\newcommand{\cy}[1]{\authorcomment{blue}{Chaoyi:} \textcolor{purple}{\textit{#1}}}
\newcommand{\lc}[1]{\authorcomment{red}{Lydia:} \textcolor{red}{\textit{#1}}}
\newcommand{\je}[1]{\authorcomment{pink}{Jeroen:} \textcolor{pink}{\textit{#1}}}

\else
\newcommand{\authorcomment}[2]{}

\newcommand{\cy}[1]{}
\newcommand{\lc}[1]{}
\newcommand{\je}[1]{}
\fi

\begin{document}
\title{
    \includegraphics[width=8cm, keepaspectratio]{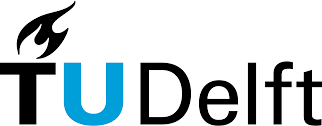}\\
    \vspace*{2cm}
    \textbf{
        \textless Watermarking Diffusion Graph Models\textgreater\\
        {\large \textless   \texttt{GUISE}: Graph GaUssIan Shading watErmark\textgreater
        }
    }
    \vspace*{1cm}
}

\author{
    \textless \textbf{
Renyi Yang$^1$\textgreater}\\
    \hfill \break
    \textbf{Supervisor(s): \textless Lydia Chen$^1$\textgreater, \textless Jeroen Galjaard $^1$\textgreater \textless Chaoyi Zhu$^1$\textgreater}\\
    \break
    {\large 
        \hfill \break
        $^1$EEMCS, Delft University of Technology, The Netherlands
    }\\
}

\date{}

\maketitle
\thispagestyle{empty}

\let\clearpagebackup\clearpage
\renewcommand{\clearpage}{ }

\onecolumn

\vspace*{1.5cm}
\begin{center}
    A Thesis Submitted to EEMCS Faculty Delft University of Technology,\\
    In Partial Fulfilment of the Requirements\\
    For the Bachelor of Computer Science and Engineering\\
    \today
\end{center}

\vspace*{2cm}

\noindent
{\small
Name of the student: \textless Renyi Yang\textgreater\\
Final project course: CSE3000 Research Project\\
Thesis committee: \textless Lydia Chen \textgreater, \textless Chaoyi Zhu \textgreater, \textless Jeroen Galjaard \textgreater 
\textless Rihan Hai\textgreater \\
}
\vfill

\begin{center}
    An electronic version of this thesis is available at http://repository.tudelft.nl/.
\end{center}

\twocolumn
\let\clearpage\clearpagebackup  
\clearpage
\setcounter{page}{1}

\begin{abstract}

In the expanding field of generative artificial intelligence, the integration of robust watermarking technologies is essential to protect intellectual property and maintain content authenticity. Traditionally, watermarking techniques have been developed primarily for rich information media such as images and audio. However, these methods have not been adequately adapted for graph-based data, particularly on molecular graphs. Latent 3D graph diffusion(LDM-3DG)~\cite{you2023latent} is an ascendant approach in the molecular graph generation field. This model effectively manages the complexities of molecular structures, preserving essential symmetries and topological features. To protect this sophisticated new technology, we adapt the Gaussian Shading~\cite{yang2024gaussian}, a proven performance lossless watermarking technique, to the latent graph diffusion domain. 
Our adaptation simplifies the watermark diffusion process through duplication and padding, making it adaptable and suitable for various message types. 
We conduct several experiments using the LDM-3DG model on publicly available datasets QM9~\cite{ramakrishnan2014quantum} and Drugs~\cite{axelrod2022geom}, to assess the robustness and effectiveness of our technique. Our results demonstrate that the watermarked molecules maintain statistical parity in 9 out of 10 performance metrics compared to the original. Moreover, they exhibit a 100\% detection rate and a 99\% extraction rate in a 2D decoded pipeline, while also showing robustness against post-editing attacks.


\end{abstract}
\section{Introduction}

Diffusion models are extensively used to generate graphs, playing a pivotal role in various domains~\cite{zhang2023survey}. In molecular generation, these models are pioneering state-of-the-art advancements, addressing diverse tasks from molecule property prediction to structure-guided protein design. By enabling more efficient exploration of molecular interactions, diffusion models can accelerate the identification of promising drug candidates, thereby revolutionizing the pharmaceutical industry. Capitalizing on these advantages, the Latent 3D Graph Diffusion (LDM-3DG)~\cite{you2023latent} model leverages diffusion generative models to effectively capture the complex distributions of 3D graphs. It achieves this by diffusing in a low-dimensional latent space, which not only enhances the ability to handle molecular complexities but also maintains essential symmetries and topological features. This model utilizes cascaded 2D-3D graph autoencoders to learn this latent space, which not only reduces reconstruction errors but also maintains symmetry group invariance. This approach not only improves the quality of generation but also significantly speeds up training, demonstrating great potential in applications such as drug discovery.


The absence of robust safeguards for diffusion model-generated data can lead to substantial challenges, including misinformation and conflicts over intellectual property. Ensuring the traceability and accountability of these models is essential for two main reasons. First, a powerful diffusion model requires substantial computational resources and meticulously annotated data for its creation, it must be protected from being exploited by unauthorized entities offering paid services. Second, unverified Artificial Intelligence(AI) generated content can spread misinformation, undermining public trust and potentially causing harm. For instance, in the pharmaceutical industry, the unauthorized use of those generated molecular designs could result in untested and potentially harmful compounds.

Addressing these issues requires a focused approach to authenticating the source of content generated by diffusion models. Prior research has extensively explored watermarking for diffusion models, mainly focusing on multimedia carriers such as images~\cite{wen2023tree} and audio~\cite{cao2023invisible} where abundant information allows for effective watermark embedding. These studies have demonstrated the feasibility of embedding watermarks to protect intellectual property and verify authenticity. However, these techniques have not been adequately adapted for graph-based data which often contains less overt data per unit and operates in smaller latent spaces. This project seeks to fill this gap by developing effective watermarking methods tailored to molecular graphs, ensuring the integrity and traceability of graphs generated by diffusion models.

We address this issue by adapting Gaussian Shading—a technique originally developed for image diffusion models—to graph-based models. Our adaptation includes adding padding bits after watermark duplication to handle that 500 latent space length of LDM-3DG is not a multiple of bytes(8 bits).
Furthermore, Instead of using the fixed watermark capacity in the original Gaussian Shading approach, we simplify this process by directly replicating the watermark message. According to ablation studies from Gaussian Shading~\cite{yang2024gaussian}, the length of the watermark capacity minimally impacts detection rates. The adaptations allow our method to effectively embed watermark messages of various lengths and make the technique flexible for different shapes and sizes of latent spaces, demonstrating its potential for robust watermark protection across various data modalities.

This research focuses on developing small-molecule generative models using the latent diffusion model architecture. These models are lightweight and should be suitable for deployment on personal computers. The datasets employed, QM9~\cite{ramakrishnan2014quantum} and Drugs~\cite{axelrod2022geom} from Nature publications, are publicly accessible and widely utilized in computational chemistry and drug discovery.

The main contributions of our work are:

\begin{enumerate}
    \item We successfully adapt Gaussian Shading to the LDM-3DG molecular graph diffusion model, demonstrating the technique's applicability beyond its initial image-focused context.
     \item We benchmark the performance of the generated molecules using various metrics, which demonstrate that the watermarking does not impair the model’s performance while achieving a high detection rate.
    \item We pioneer unique attack methods on watermarks in the domain of graph-based molecular generation and prove the robustness of our watermarking against modifications to the generated data.

\end{enumerate}

\section{Related Work}

\textbf{Graph diffusion models} \quad  Diffusion models have become a critical component in the field of graph generation deep learning, offering a robust framework to facilitate the spread of information across a graph's nodes and edges. Diffusion models enhance the scalability and efficiency of graph neural networks, especially for managing large-scale graphs~\cite{ingraham2019generative}. Additionally, they integrate global information into the learning process, which leads to enhanced precision in predictions and improves overall performance~\cite{atwood2016diffusion}.

\begin{figure}
  \centering
  \includegraphics[width=0.5\textwidth]{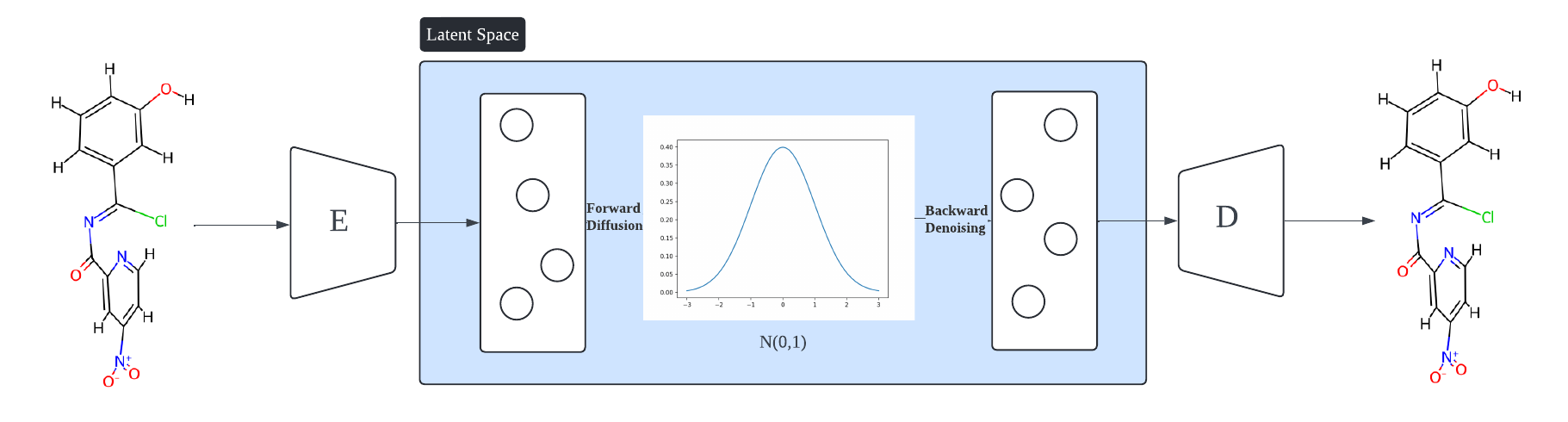}
  \caption{Architecture of the latent diffusion model. The left half represents the training phase where data is encoded to latent space and diffused to Gaussian noise. The right half represents the inference phase where noise is denoised to latent representation and decoded to data. }\label{fig:LDM}
\end{figure}

Building upon the foundational concepts of graph diffusion models, traditional discrete diffusion approaches such as DiGress meticulously modify graph structures step-by-step, allowing for precise control during generation~\cite{vignac2022digress}. This architecture, while detailed, often requires complex computation. Building on discrete diffusion models, Latent Diffusion Models (LDM) offer a streamlined approach[\ref{fig:LDM}] by encoding graph structures into a compact latent space. This latent representation undergoes a controlled diffusion process, simplifying the graph's complex properties before reconstructing it into its original form.

This method underpins innovative architectures such as EDM and NGG, EDM~\cite{guan20233d} specifically addresses the generation of 3D molecular structures with physical accuracy by accommodating SE(3) 
transformations, which presents a rigid body motion in three-dimensional space, consisting of a rotation and a translation. It ensures that generated molecules are not only structurally accurate but also physically plausible. Conversely, NGG exploits a variational autoencoder~\cite{evdaimon2024neural}
alongside a latent space diffusion process to adeptly manage the conditioning on specific graph properties, demonstrating not only efficiency but also adaptability in generating detailed graphs with desired attributes.


\textbf{Watermarking diffusion models} \quad Digital watermarking serves as a powerful tool for copyright protection and content verification by allowing the embedding of copyright or traceable identifiers directly into the data. This technique ensures that ownership information is seamlessly integrated with the content, making it easier to track and manage rights. Watermarks for diffusion models can be categorized into two primary approaches: embedding watermarks in the data and fine-tuning the model. 

 Watermark Diffusion Model (WDM)~\cite{peng2023protecting} and Stable Signature~\cite{fernandez2023stable} are typical approaches to fine-tuning the model. WDM incorporates a distinct Watermark Diffusion Process (WDP) into the standard diffusion process used for generative tasks. During the embedding phase, the model is trained such that it modifies the standard diffusion process to include characteristics specific to the watermark, using a special optimization objective that ensures the watermark's properties are preserved. Stable Signature embedding the watermark by fine-tuning the decoder, evaluations confirm that the method effectively identifies the source of generated images, maintaining high detection accuracy even when images are cropped or undergo other transformations.

For data embedding methods there is DiffusionShield~\cite{cui2023diffusionshield} which uses ``pattern uniformity'' to embed consistent watermark patterns throughout the training dataset. The watermarks are divided into basic patches that represent binary messages and are uniformly applied, which improves detection accuracy and reduces image distortion. And Tree Ring~\cite{wen2023tree} embeds the watermark in the initial latent noise during sampling, This watermark is constructed in Fourier space, and the process of extracting the watermark involves a reverse diffusion step, and then checking whether it contains the embedded signal.

Fine-tuning-based watermarking methods require retraining the model, which can be resource-intensive and inflexible. Data embedding methods typically alter the data format, which can affect model performance. There is a critical need for a more lightweight, low-overhead, plug-and-play method that does not compromise model performance and can be easily distributed and utilized by individuals. 
In response to these challenges, the method of Gaussian Shading~\cite{yang2024gaussian} emerges as a promising solution. We opt for Gaussian Shading (GS) over Tree Ring due to its theoretically lossless nature and its broader applicability across various latent space dimensions. Unlike Tree Ring, which requires a two-dimensional or higher latent space, GS naturally adapts to any shape or size of the latent space.


 \begin{figure*}[h]
  \centering
  \includegraphics[width=\textwidth]{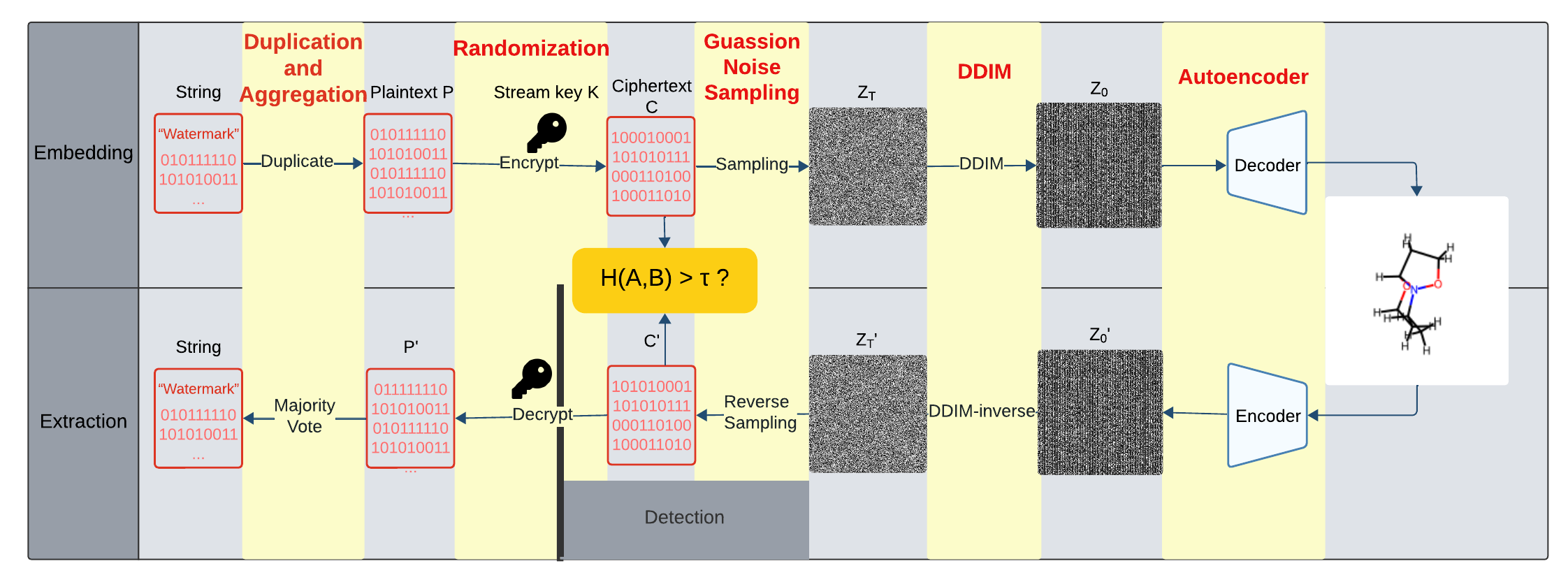} 
  \caption{\algo Framework. the watermark is duplicated and encrypted to generate a random bitstream, then we use Gaussian noise sampling to generate the 
 latent, DDIM-sampling, and decode it to create a watermarked molecule. The watermark is extracted by reversing these operations and detected by comparing Hamming distances between bitstreams.}\label{fig:architecture}
\end{figure*}
\section{\algo Framework}

In this research, 
we develop an algorithmic framework, 
which feature two primary functionalities: embedding and detection. During the embedding process, a string $msg$ which represents the watermark message is integrated into the graph generation model. This integration ensures that the graphs sampled from the model inherently carry the information contained in $msg$ 
For the detection process, the objective is to accurately identify the presence of the watermark $msg$ within the generated graphs. Additionally, we address a relaxed version of this problem, where given the watermark message $msg$, the algorithm determines whether a generated graph contains the embedded $msg$ without explicitly reconstructing the message.

We adapt Gaussian shading~\cite{yang2024gaussian} as a solution to this problem.
 The overall watermark architecture architecture is illustrated in Figure~\ref{fig:architecture}. Gaussian shading utilizes a pseudo-random algorithm to generate Gaussian noise. This noise corresponds bit-by-bit with each bit of the watermark message m within the latent space representation. This noise is then processed through a Denoising Diffusion Implicit Model (DDIM) 
~\cite{song2020denoising} 
  and a decoder to create a graph representation, thereby embedding the watermark into the graph. The embedding algorithm is provided in Algorithm~\ref{alg:watermark_embedding}

Subsections 3.1 to 3.4 provide a detailed description of the watermarking process, 3.5 provides and its inverse(extraction) process and detection process.

\begin{algorithm}
\caption{Watermark Embedding Algorithm}
\label{alg:watermark_embedding}
\begin{algorithmic}[1]
\REQUIRE Message $m$, Key $k$, Nonce $n$
\ENSURE Watermarked latent variables $z$

\STATE Repeat the message $m$ to form $m'$, where $|m'| = 500$ bits.
\STATE Let $m'_b$ be the byte representation of $m'$.
\STATE Encrypt $m'_b$ using ChaCha20 to obtain the ciphertext $c$: 
\[
c = \text{ChaCha20}(m'_b, k, n)
\]

\STATE Initialize $l \leftarrow 1$ and an empty list $z$.
\FOR{each bit $y$ in $c$}
    \STATE Generate $u \sim U(0, 1)$.
    \STATE Compute $z_i = \text{ppf}\left(\frac{y + u}{2^l}\right)$.
    \STATE Append $z_i$ to $z$.
\ENDFOR
\RETURN $z$
\end{algorithmic}
\end{algorithm}

\subsection{Watermark Duplication and Aggregation}
To address the disparity between the fixed dimensionality of the model's latent space, \( n \) dimensions, and the variable length of the watermark, \( m \) bits, the watermark message is extended to match the latent space's capacity. This is done through a process of replication followed by padding with zeros. The diffusion process, which adapts an \( m \)-bit message into an \( n \)-bit message, can be expressed in Eq.~\ref{eq:watermark_diffusion}, where \( \oplus \) denotes concatenation.

\begin{equation}
\text{P} = (\underbrace{m, m, \ldots, m}_{\text{\textit{n/m} times}}) \oplus (\underbrace{0, \ldots, 0}_{\text{\textit{n\%m} times}})
\label{eq:watermark_diffusion}
\end{equation}

The aggregation process recovers the original \( m \)-bit watermark from the extended \( n \)-bit version. This is achieved by computing a weighted majority for each bit in the original message based on its replicas. The reconstructed bit \( m_i \) is calculated using Eq.~\ref{eq:watermark_aggregation},
\begin{equation}
m_i = \left\lfloor \frac{1}{x} \sum_{j=1}^{x} v_{ij}\right\rfloor,
\label{eq:watermark_aggregation}
\end{equation}
where \( x = \lfloor \frac{n}{m} \rfloor  \) is the number of replicas per original bit, \( v_{ij} \) represents the value of the \( j \)-th replica for the \( i \)-th bit. This aggregation ensures that the watermark is seamlessly integrated and retrievable despite variations in length and dimensionality constraints. The voting mechanism makes the watermark detection method resistant to the loss of information in the DDIM-inverse and encoding process.

\subsection{Watermark Randomization}
After the process of watermark diffusion, we obtain a plaintext bitstream $P$. Our next objective is to randomize this bit sequence so that each bit has an equal probability of being 0 or 1. 
This randomization facilitates the subsequent generation of Gaussian noise. If the original message is not encrypted to achieve a uniform distribution, direct processing through Gaussion noise sampling may result in matrices that do not conform to a Gaussian distribution. Watermark data concentrated in specific areas can produce perceptible noise or irregularities in the generated content, adversely affecting the generation quality. Furthermore, non-uniform distribution of watermark data could create detectable statistical features in the generated graphs, making them vulnerable to discovery and removal by attackers.

The chosen method for achieving this is the use of the ChaCha20~\cite{bernstein2008chacha} stream cipher. ChaCha20 is an encryption algorithm that enhances security by thoroughly mixing input bits using additions, XORs, and rotations across multiple rounds. It builds upon the design principles of its predecessor, Salsa20, but with improved dispersion
per round, making it resistant to cryptanalysis while maintaining low-overhead. The watermark extraction process involves decryption. If the key and nonce are the same, the ChaCha20 encryption and decryption processes are deterministic and invertible. This ensures that the original bitstream $P$ can be accurately recovered during the extraction phase.



\subsection{Guassion Noise Sampling}
To further generate the Gaussian noise latent from ciphertext $C$, we use Eq.~\ref{eq:embedding} for sampling,
\begin{equation}
z_i = \text{ppf}\left(\frac{b_i + u}{2}\right),
\label{eq:embedding}
\end{equation}
where \( u \sim U(0,1) \) is a uniformly chosen random variable, \( b_i \) is the \(i\)-th bit of the ciphertext, and \( z_i \) is the corresponding Gaussian-distributed latent variable in latent space. Specifically, we divide the Gaussian range into two regions: one for 0 bits and the other for 1 bits. 0 bits correspond to negative values and 1 bits correspond to positive values. The ppf function maps the cumulative probability to the value of the Gaussian variable. Since the cumulative probability is uniformly distributed between 0 and 1, we only need to ensure that the parameter of ppf is randomly distributed within range (0, 1) to guarantee that our sampling result follows a Gaussian distribution. The transformation \(\frac{b_i + u}{2}\) precisely converts the discrete distribution of 0 and 1 into a continuous distribution within (0, 1). Therefore, our watermark embedding process is lossless

To get ciphertext from latent, we perform reverse sampling in Eq.~\ref{eq:extracting},
\begin{equation}\label{eq:extracting}
b_i = \left\lfloor 2 \cdot \text{cdf}(z_i) \right\rfloor,
\end{equation}
where the cdf function converts the latent Gaussian value \( z_i \) back into the bit \( b_i \). 
Moreover, the Gaussian noise regions can also be divided into \( N \) parts instead of just two, where each part corresponds to \(\log_2(N)\) bits in the ciphertext. This provides a more granular approach to embedding and extracting bits.

\begin{table*}[tbp]
\centering
\caption{Comparison of molecular generation performance between original and watermarked models on QM9 and GEOM-drugs datasets. Valid: proportion of (POF) chemically valid molecules; Valid\&Uni: POF chemically valid and unique molecules; AtomSta: POF atoms with correct valency; MolSta: POF molecules without unstable atoms. The higher the better. t-statistic: t-value under the hypothesis of equality mean value}\label{tab:performance_comparison}
\begin{tabular}{@{}lcccccccc@{}}
\toprule
\multicolumn{1}{c}{\textbf{Methods}} & \multicolumn{4}{c}{\textbf{QM9}} & \multicolumn{4}{c}{\textbf{GEOM-drugs}} \\ \cmidrule(l){2-5} \cmidrule(l){6-9} 
 & \textbf{Valid} $\uparrow$ & \textbf{Valid\&Uni} $\uparrow$ & \textbf{AtomSta} $\uparrow$ & \textbf{MolSta} $\uparrow$ & \textbf{Valid} $\uparrow$ & \textbf{Valid\&Uni} $\uparrow$ & \textbf{AtomSta} $\uparrow$ & \textbf{MolSta} $\uparrow$ \\ \midrule
Original     & 1.00(0) & 97.83(0.04) & 94.5(0.20) & 81.01(0.30) & 1.00(0) & 99.99(0) & 79.75(0.11) & 4.21(0.27) \\
Watermarked  & 1.00(0) & 98.10(0.22) & 94.41(0.27) & 80.84(0.24) & 1.00(0) & 1.00(0) & 79.64(0.11) & 4.23(0.40) \\ 
t-statistic  & - & 2.09 & 0.46 & 0.77 & - & - & 1.23 & 0.07 \\ \bottomrule
\end{tabular}
\end{table*}

\begin{table*}[tbp]
\centering
\caption{Distribution discrepancy metrics between original and watermarked models on QM9. MW: Molecular weight; ALogP: Octanol-water partition coefficient; PSA: Polar surface area; QED: Drug likeness; FCD: Frechet ChemNet distance; Energy: Conformer energy, Hartree as unit. Other metrics represent total variation distances (×1e-2) of certain molecular properties. The lower the better. t-statistic: t-value under the hypothesis of equality mean value}
\label{tab:metrics_comparison}
\begin{tabular}{@{}lcccccc@{}}
\toprule
\textbf{Methods} & \textbf{MW} $\downarrow$ & \textbf{ALogP} $\downarrow$ & \textbf{PSA} $\downarrow$ & \textbf{QED} $\downarrow$ & \textbf{FCD} $\downarrow$ & \textbf{Energy} $\downarrow$ \\ \midrule
Original   & 4.30(0.28) & 1.84(0.25) & 2.19(0.35) & 2.16(0.06) & 147.15(5.09) & 3.36(0.50) \\
Watermarked & 4.38(0.29) &1.88(0.38) & 2.65(0.22) & 2.43(0.25) & 180.65(8.57)  & 3.15(0.24) \\
t-statistic & 0.34 & 0.15 & 1.93 & 1.58 & \textcolor{red}{5.82}  & 0.66 \\ \bottomrule
\end{tabular}
\end{table*}

\subsection{DDIM and Autoencoder}
After sampling, the random latent \( Z_t \) is transformed into the latent representation \( Z_0 \) using DDIM. DDIM (Denoising Diffusion Implicit Models) is a method that iteratively denoises the latent variables, effectively transforming them into a more stable representation~\cite{song2020denoising} The reason for choosing DDIM over DDPM sampling is that DDIM is deterministic and can be reconstructed. The DDIM process can be described in Equation~\ref{eq:forward}, where \( z_t \) represents the state at time \( t \), \( \alpha \) are redefined variance scaling coefficients, \( \theta^{(t)}(z_t) \) is the noise estimation model at time \( t \) given the state \( z_t \), \( \Delta t \) represents the time step interval. \( \Delta t \) and \( \alpha \) are affected by the scheduling policy and sampling steps, Common scheduling strategies include Linear, Cosine, Quadratic, or Exponential.

\begin{multline}
z_{t-\Delta t} = \frac{1}{\sqrt{\alpha_{t-\Delta t}}} \left( z_t \sqrt{\alpha_t} \right. \\ 
\left. + \frac{1}{2} \left(\sqrt{\frac{1-\alpha_t}{\alpha_t}} - \sqrt{\frac{1-\alpha_{t-\Delta t}}{\alpha_{t-\Delta t}}}\right) \theta^{(t)}(z_t) \right)
\label{eq:forward}
\end{multline}

\begin{multline}
z_{t+\Delta t} = \frac{1}{\sqrt{\alpha_{t+\Delta t}}} \left( z_t \sqrt{\alpha_t} \right. \\
\left. + \frac{1}{2} \left(\sqrt{\frac{1-\alpha_t}{\alpha_t}} - \sqrt{\frac{1-\alpha_{t+\Delta t}}{\alpha_{t+\Delta t}}}\right) \theta^{(t)}(z_t) \right)
\label{eq:reverse}
\end{multline}

The Autoencoder then translates this latent representation into the graph. Different architectures may employ various types of encoders; Many graph diffusion models employ Variational Autoencoders (VAE), a generative model that learns to encode and decode data while capturing the underlying data distribution, and it uses Graph Neural Networks (GNN)~\cite{micheli2009neural} as the backbone for the encoder and decoder. In autoencoders, the encoding and decoding stages are designed to be reciprocal, this symmetry is achieved through training the network to minimize the reconstruction error.

\begin{algorithm}
\caption{Watermark Detection Algorithm}
\label{alg:watermark_detection}
\begin{algorithmic}[1]
\REQUIRE Latent variables $z$, Key $k$, Nonce $n$, Threshold $\tau$, Message $m$
\ENSURE Boolean detection result

\STATE Compute the bits $c_i$ from the latent variables $z$:
\[
c_i = \left\lfloor 2^l \cdot \text{cdf}(z_i) \right\rfloor
\]

\STATE Compute secondary ciphertext $c'$ using Algorithm 1 steps 1-3:
\[
 c' = \text{Algorithm 1}_{1-3}(m, k, n)
\]

\STATE Determine if the watermark m is contained using the Hamming distance:
\[
\text{If\_detected} = (H(c, c') \leq \tau)
\]

\RETURN \text{If\_detected}
\end{algorithmic}
\end{algorithm}
\begin{algorithm}
\caption{Watermark Extraction Algorithm}
\label{alg:watermark_extraction}
\begin{algorithmic}[1]
\REQUIRE Latent variables $z$, Key $k$, Nonce $n$
\ENSURE Extracted message string $m$
\STATE Compute the bits $c_i$ from the latent variables $z$:
\[
c_i = \left\lfloor 2^l \cdot \text{cdf}(z_i) \right\rfloor
\]

\STATE Convert the bit array $c$ to bytes and decrypt using ChaCha20:
\[
p = \text{decrypt}(\text{packbits}(c), k, n)
\]

\STATE Apply majority voting across each n bits of $p$ to obtain the final extracted message bits $m$:
\[
m = \text{majority\_vote}(p)
\]

\RETURN $m$
\end{algorithmic}
\end{algorithm}

\subsection{Detection and Extraction}
Watermark Detection and Extraction first involves encoding the graph into its latent representation, followed by applying the DDIM-inverse to obtain \( Z_0 \) using Equation~\ref{eq:reverse}. We then employ reverse sampling on the latent vector \(z\) to generate the ciphertext \(c\), as explained in Section 3.3.


For the detection of watermarks as described in Algorithm~\ref{alg:watermark_detection}, since the message \(m\) is assumed to be known, the goal is to verify whether the generated output contains the watermark of \(m\). We re-encrypt \(m\) according to steps 1-3 of Algorithm~\ref{alg:watermark_embedding} to produce a bitstream \(c'\). We then measure the Hamming distance between \(c\) and \(c'\). If this distance is no more than a specified threshold \(\tau\), it is inferred that the graph contains the watermark. The relationship between \(\tau\) and the false positive rate (FPR) is analyzed to demonstrate the theoretical viability of this method, as detailed in Appendix A.

For extraction as described in Algorithm~\ref{alg:watermark_extraction}, we decrypt the ciphertext \(c\) to obtain plaintext \(p\), and then apply the aggregation method outlined in Section 3.1 to implement majority voting to determine each bit of \(m\).

We conclude that the technique introduced by Guassion Shading achieves 100\% detection and extraction accuracy because Algorithm~\ref{alg:watermark_embedding} and Algorithm~\ref{alg:watermark_extraction} are deterministic and mutually reversible. 
However, these two algorithms cannot function independently as they rely on the graph sampling method and its reverse sampling. The efficacy of Gaussian Shading is rather inherently dependent on the performance of the following components and assumptions within the latent diffusion model:
\begin{enumerate}
    \item The reconstruction capability of the autoencoder, so that \( Z_0 \approx \text{Enc}(\text{Dec}(Z_0)) \).
    \item The denoising performance of the diffusion model, so that $Z_t' \sim \mathcal{N}(0, 1)$. If $Z_t'$ is scaled or shifted, ciphertexts $c$ will be biased and lead to a higher bit error.
    \item The reversibility of sampling methodology of the diffusion model, so that \( Z_t' \approx Z_t \).
\end{enumerate}

Section~\ref{subsection4.4} conducts experiments about how inversion quality affects the performance of Gaussian shading.


\section{Experimental Results}
Here we present the experimental analysis to assess the impact of watermarking on the quality of graph generation.
First, we outline the experimental setup employed. 
Following, we evaluate the performance of models that incorporate watermarks relative to their unmarked counterparts. 
We then simulate various attacks on the generated graphs to ascertain the robustness of the watermark detection mechanism under diverse conditions. 
\subsection{Experimental Setting}
In this paper, we explore watermarking techniques within the context of 3D molecular graph generation, utilizing the LDM-3DG model~\cite{you2023latent}. 
LDM-3DG is an innovative latent diffusion model for generating molecular structures, distinguished by its approach of separately handling the topology (connectivities) and geometry (spatial coordinates) of molecules. This is achieved through two cascaded decoders. The model's autoencoder component consists of a 2D encoder employing a Variational Autoencoder with a Hierarchical Message Passing Network (HierMPN) architecture, and a 3D encoder using a Graph Neural Network (GNN) based on Multilayer Perceptrons (MLP). The autoencoder is trained on large-scale public databases such as ChEMBL~\cite{gaulton2012chembl} and PubChemQC~\cite{nakata2017pubchemqc}.

We perform 500 steps of unconditional sampling using DDIM and DDIM-inverse. Gaussian shading was employed with the window value l=1. All experiments are conducted using the PyTorch 1.13.1
framework, running on a single RTX 4080 GPU and a Ubuntu 22.04 operation system.

\subsection{Model Performance}
We conduct performance tests on two distinct datasets to evaluate the performance of the model after our watermarking method in 3D molecular graph generation. The first dataset, QM9~\cite{ramakrishnan2014quantum}, comprises approximately 134,000 small organic molecules with up to nine heavy atoms. The second dataset, GEOM-drugs~\cite{axelrod2022geom}, contains 450K larger molecules up to 181 atoms, and it's typically found in drug discovery datasets. Those datasets are computationally feasible for chemistry calculations and are extensively used for benchmarking molecular generation models.

Our evaluation metrics include the quality of generated molecules and the distribution discrepancy. The quality metric assesses the chemical validity and stability of the generated molecules, ensuring that they comply with established chemical rules and possess realistic molecular configurations 
The distribution discrepancy metric measures how closely the properties of generated molecules match those of real molecules in the test dataset, considering aspects like molecular weight, polarity, and pharmacological potential. Detailed explanation of all matrics used in this section and it's generation methods are in Appendix~\ref{app:C}.

We benchmark molecules generated under two conditions: with and without the ``Watermark" string embedded. We perform a 3-fold test using a set of 30,000 generated molecules to ensure the reliability and consistency of our evaluation. We subsequently conduct t-tests on all the metrics, assuming a null hypothesis (\(H_0\)) of no difference between the original distribution and watermarked distribution and an alternative hypothesis (\(H_a\)) that a significant difference exists. We calculate the t-statistic for each matric. The results are contained in Table \ref{tab:performance_comparison} \ref{tab:metrics_comparison}

Given that we use 3-fold validation and set the significance level \(\alpha = 0.05\), the critical threshold from the t-distribution should be \(t_{0.05}(4) = 2.13\). We accept the null hypothesis if the computed t-value is less than 2.13. For all metrics except FCD, all metrics adhere to the null hypothesis (\(H_0\)) of no significant difference between the original and watermarked distributions. This implies that watermarking does not adversely affect the molecular generation model’s performance statistically. Notably, the model showcases robust performance across both QM9 and GEOM-drugs datasets, suggesting that our watermarking method is effectively generalizable. It should be noted that since both the watermarked and non-watermarked models achieved perfect performance (100\%) on some indicators, the statistical tests performed on these indicators are not of practical significance.

In terms of distribution discrepancy, apart from the Frechet ChemNet Distance (FCD), no significant differences were found in other metrics, which aligns with the theoretical expectation that Gaussian shading should be lossless. The significant variance observed in FCD may suggest that the model has not yet converged with the sample size provided, resulting in greater fluctuation.

\subsection{Watermark Robustness}
\begin{figure}[ht]
    \centering
    \includegraphics[width=0.5\textwidth]{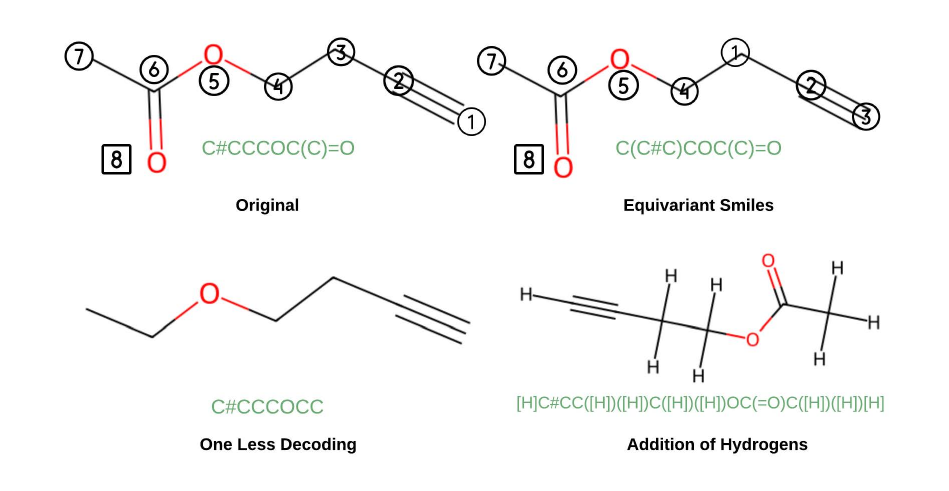}
    \caption{Illustation of molecule ``C\#CCCOC(C)=O" and it's Equivariant Smiles, Addition of Hydrogens and One Less Decoding representation}
    \label{fig:attacks}
\end{figure}

In this section, we assess the robustness of our graph watermarking scheme against various attacks, which notably involve common modifications to the molecular topology. Specifically, we conduct these attacks by modifying the SMILES (Simplified Molecular Input Line Entry System) representations of the molecules. SMILES is a notation that allows a user to represent a chemical structure in a way that can be used by the computer. It encodes the structure of a molecule using short ASCII strings. For example, the SMILES representation of water is ``O", while ethanol is represented as ``CCO". This system is widely used for its simplicity and ability to efficiently describe complex molecular structures~\cite{weininger1988smiles}. 
The three attacks are illustrated in Figure~\ref{fig:attacks}.

\textbf{Equivariant Smiles} \quad  SMILES enumeration is a technique that takes advantage of the fact that a single molecule can be represented by multiple different SMILES strings. This method involves generating multiple SMILES representations for the same molecule, thereby augmenting the dataset and potentially improving the robustness and performance of machine-learning models-\cite{bjerrum2017smiles}. We utilize this technique to modify the SMILES generated by the LDM-3DG model. For example, the SMILES string ``CCCC(CO)CO" can be changed to ``C(C(CO)CO)CC". Both SMILES represent the same molecular structure from a chemical perspective, but the different representations arise due to variations in the starting atom and the order of atom connections.

\textbf{Addition of Hydrogens} \quad  In most organic structures represented by SMILES, hydrogen atoms are typically implied rather than explicitly noted, adhering to the normal valence rules without specific mention in the SMILES string. we explicitly specify hydrogen atoms within the molecular structure. For instance, a simple molecule like ethane, normally represented as ``CC" in SMILES, is modified to ``C([H3])C([H3])" to explicitly show all hydrogen atoms attached to each carbon. This alteration tests our watermarking system's capability to handle and recover watermarks from these denser, more detailed molecular descriptions.

\textbf{One Less Decoding} \quad 
In the context of the LDM-3DG's iterative decoder structure, our approach exploits the inherent step-by-step motif-based graph generation. During the decoding process, the model utilizes a depth-first search (DFS) mechanism, where each motif is selected and expanded in sequence. This hierarchical, iterative 
generation is crucial as it allows the integration of complex molecular structures sequentially, ensuring that each part of the molecule is correctly positioned in relation to its predecessors~\cite{jin2020hierarchical}.
In our specific attack method, we capitalize on this sequential generation by intercepting and storing intermediate molecular graphs at various stages of the decoding process. Instead of completing the full sequence of motif additions, we halt the process prematurely, thus obtaining a molecular structure that is one step less decoded than the final intended structure. For example, if the fully decoded structure is ``C\#CCCOC(C)=O". Halting one step earlier could yield ``C\#CCCOCC".
This strategy tests the watermark’s resilience not only to modifications of the molecule's intended structure but also evaluates how well the watermark can be preserved and recognized in partially generated or interrupted synthesis scenarios.

\begin{figure}[ht]
    \centering
    \includegraphics[width=0.4\textwidth]{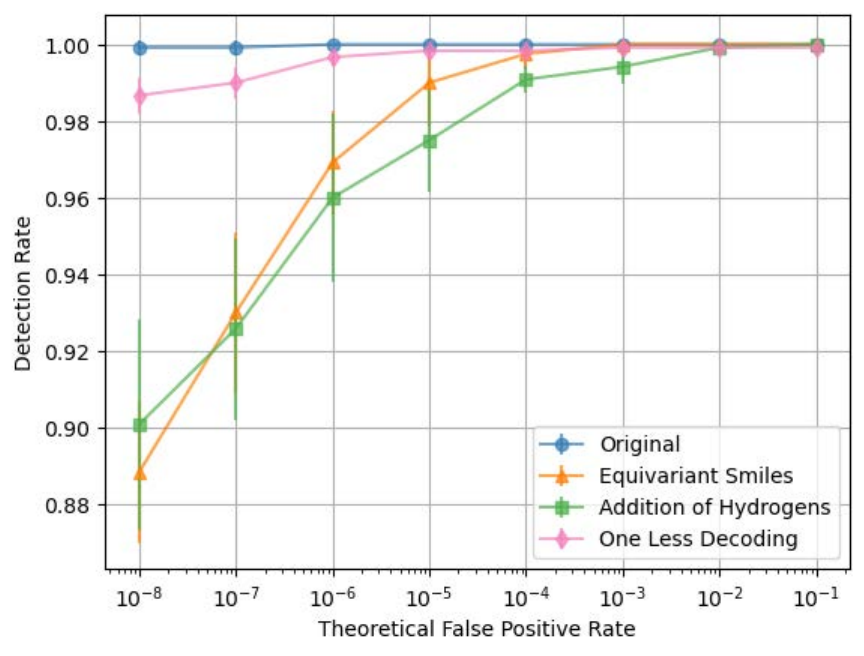}
    \caption{The watermark detection rate against theoretical false positive rate under the watermarked molecule set and three modified molecule sets}
    \label{fig:detection_rate}
\end{figure}

The robustness results are contained in Figure~\ref{fig:detection_rate}. We create a dataset of 1,200 molecules, each embedded with the string ``watermark" as a watermark. To test the robustness of the watermark against structural perturbations, we apply the previously described attacks—Equivariant SMILES, Addition of Hydrogens, and One Less Decoding—to this dataset. Each attack generated an additional 1,200 modified molecules, resulting in a total of 4,800 watermarked molecules.

We employ a 6-fold cross-validation method, each fold consists of 200 randomly selected molecules from each dataset. We vary the theoretical false positive rate(TFPR) to calculate the detection rates for the watermark across these four groups (original and attacked watermarked molecules). The relationship between threshold and TFPR is $threshold = binom.ppf(TFPR) \quad   binom \sim Bin(n,0.5)$ where n is the length of the latent space.
To assess the statistical reliability of our watermark detection rates, we compute the confidence intervals around the mean detection rates for each set of molecules. These intervals are determined using a standard deviation-based approach, typically denotes as $\mu \pm \sigma$ where $\mu$ represents the mean and $\sigma$ represents the standard deviation of the detection rates.  

\begin{figure}[ht]
    \centering
    \includegraphics[width=0.4\textwidth]{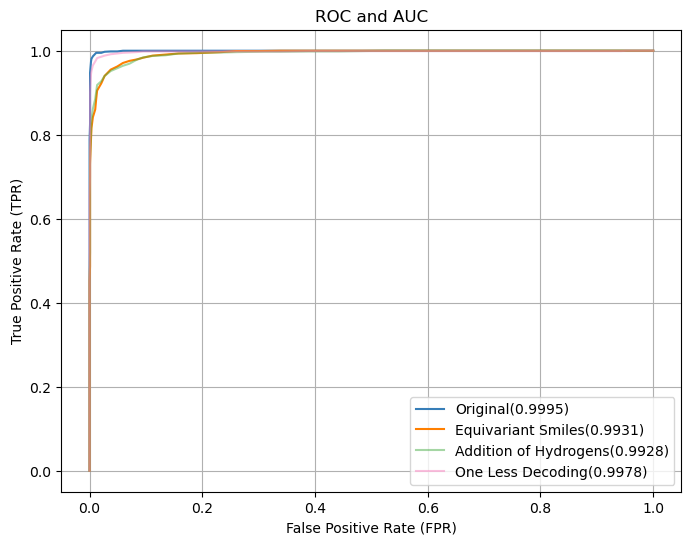}
    \caption{The detection rate of watermarked and modified datasets against the detection rate of none watermarked datasets, Area under the ROC Curve is reported in legends}
    \label{fig:ROC}
\end{figure}

For comparative analysis, we also include a control group of 1,200 not watermarked, regular molecules to calculate the experimental false positive rate(EFPR) we construct a Receiver Operating Characteristic (ROC) curve and calculate AUC in Figure~\ref{fig:ROC}.

Despite the detection rate being negatively affected due to the attacks, the watermark detection rates remained distinctly higher compared to the control group(not watermarked molecules), underscoring the robustness of our watermarking approach. This substantiates the practical utility of our method in safeguarding molecular data against typical adversarial modifications.


\subsection{Inversion Quality as Impact Factor}
\label{subsection4.4}


In this section, we conduct an analysis of how the components of the latent diffusion model (LDM) affect the detection and extraction accuracy of Gaussian Shading. We design four detection pipelines that utilize different components of the LDM-3DG architecture:
\begin{enumerate}
    \item \textbf{Diffusion\_only}: We replace \( Z_t' \) with \( Z_t \). In this pipeline, only the DDIM-inverse is operational, bypassing any autoencoder, emphasizing the role of the diffusion process alone in watermark detection.
    \item \textbf{2D\_decoded}: We use \( \text{concat} \left[ Z_t'[0:250], Z_t[250:] \right] \) to replace \( Z_t' \). This pipeline solely utilizes the 2D autoencoder, extracting the watermark from the topological molecular structure, while the conformational information of the molecule in the latent space is presumed known.
    \item \textbf{3D\_decoded}: We use \( \text{concat} \left[ Z_t[0:250], Z_t'[250:] \right] \) to replace \( Z_t' \). This setup exclusively employs the 3D autoencoder, where the latent space representation of the molecular topological structure is considered known.
    \item \textbf{Fully\_decoded}: An end-to-end detection pipeline that employs all components of the LDM, retrieving watermark information directly from the 3D molecular graph.
\end{enumerate}

We embed a single-byte watermark ``X" into 800 molecular samples and evaluate the detection performance using \( Z_t \sim N(0,1) \) as the baseline. We calculate the detection rates at theoretical false positive rates (FPR) of 1\% and 5\%. We also report the extraction rate and bit accuracy. The results are presented in Table~\ref{tab:ldm_performance}.

\begin{table}[H]
\centering
\resizebox{\columnwidth}{!}{%
\begin{tabular}{@{}lcccc@{}} 
\toprule
\textbf{Method} & \textbf{TPR@1\% FPR} & \textbf{TPR@5\% FPR} & \textbf{Extraction} & \textbf{Bit Acc.} \\ 
\midrule
Diffusion only & 1.000 & 1.000 & 1.000 & 0.944 \\ 
2D decoded     & 0.999 & 1.000 & 0.990 & 0.693 \\ 
3D decoded     & 0.001 & 0.006 & 0.000 & 0.494 \\ 
Fully decoded  & 0.000 & 0.000 & 0.000 & 0.488 \\ 
\bottomrule
\end{tabular}
}
\caption{Detection rates at 0.01 and 0.05 FPR, extraction rates, and bit accuracy for four proposed pipelines}
\label{tab:ldm_performance}
\end{table}

The diffusion-only pipeline has a relatively low bit error rate, which leads to outstanding detection and extraction performance.
The 2D decoded pipeline demonstrates a high tolerance for bit errors, maintaining near-perfect detection performance (TPR at 1\% and 5\% FPR) despite introducing more bit inaccuracies (Bit Accuracy of 0.693). This indicates that the watermark detection process is robust to bit perturbations to some extent. In contrast, the 3D decoded pipeline shows significantly poorer reconstructive ability, with a bit accuracy rate close to random guessing (0.494). This low accuracy severely compromises the watermark detection, resulting in drastically low TPRs and indicating that the watermark information is effectively lost. The bad 3d autoencoder significantly affects the good quality of 2d autoencoder, marking Gaussian shading ineffective on a fully decoded pipeline. This outcome underscores the ineffectiveness of Gaussian Shading when the autoencoder's performance is inadequate.

Consequently, our research is confined to the 2D encoder. The robustness tests focused exclusively on modifications to the molecular geometric structure while disregarding the spatial structure of the molecules.

Based on a 2d encoded pipeline, we also conduct experiments on how the reversibility of the sampling method affects the detection process of Gaussian shading. We introduce two additional sampling methods, bi-directional integration approximation (BDIA)~\cite{zhang2023exact} and DPMSolver~\cite{lu2022dpm}. Compared to DDIM, DPMSolver emphasizes rapid convergence to generate high-quality samples efficiently,  whereas BDIA focuses on precise diffusion inversion through bi-directional updates for accuracy. We measure the Mean Squared Error (MSE) between \( z_T \) and its reconstructed sample through \( \text{inv\_sample}(\text{sample}(z_T)) \) to evaluate the reconstruction performance of these sampling techniques. We compute the bit accuracy as the probability of identical bits in ciphertexts \( c \) and \( c' \). The results are summarized in Table~\ref{tab:sampling_methods}.

\begin{table}[ht]
\centering
\begin{tabular}{@{}lccc@{}}
\toprule
\textbf{Method} & \textbf{DDIM} & \textbf{BDIA} & \textbf{DPMSolver} \\ 
\midrule
\textbf{Bit Acc.} & 0.70 & 0.67 & 0.67 \\
\textbf{MSE} & 0.12 & 0.40 & 0.26 \\
\bottomrule
\end{tabular}
\caption{MSE loss and bit accuracy of different sampling methods.}
\label{tab:sampling_methods}
\end{table}

Despite the variations in reconstruction performance among different sampling methods, as reflected by the MSE values, the bit accuracy remains relatively consistent. This indicates that Gaussian shading is resilient to information loss during the sampling and inverse sampling processes, and it is adaptable to various sampling techniques without significant impact on detection quality.

\section{Discussion}

We identify a significant limitation in the use of Gaussian Shading for large-scale watermarking applications, the Randomness of Gaussian shading watermarked content depends on the Randomness of key and nonce. We employ the same key and nonce for watermarking 30,000 samples, and the QM9 Validity and Uniqueness (Valid\&Uni) metric in Table~\ref{tab:performance_comparison} significantly dropped from 98.10\% to 56.71\%. 
 this concludes that when a uniform watermark message is applied across a batch of molecules, it becomes necessary to store a randomized key and nonce for each molecule to maintain uniqueness. Without this measure, the uniqueness of the model generated content is greatly compromised. 

\textbf{Responsible Research} \quad This study strictly adheres to ethical guidelines with all datasets being sourced from Nature under a CC BY license, ensuring open and accessible data use. The autoencoder and diffusion models used in this study are provided by the authors of LDM-3DG under the GPL-3.0 license. Additionally, the Gaussian Shading method and its implementation are under the MIT License. No potential conflicts of interest were identified in this study. Detailed methodologies and all original data and code are available for replication purposes on GitLab: \url{https://gitlab.ewi.tudelft.nl/dmls/courses/cse3000/watermark/graph-diffusion}.
\section{Conclusions and Future Work}
In this study, we adapt Gaussian Shading, originally designed for image diffusion models, to graph diffusion models, demonstrating its efficacy without compromising the generative performance of the original models. Our experiments on the LDM-3DG molecular generation diffusion architecture confirm the high adaptability of Gaussian Shading to LDM-based architectures. This method proves to be universally applicable across different shapes and sizes of latent spaces within the LDM framework. However, the performance of Gaussian Shading is highly contingent upon the reconstruction capabilities of the specific models, indicating that its effectiveness is directly influenced by the quality of model training.

Looking ahead, we aim to extend our testing to other graph structures, such as Erdős-Rényi random graphs and social network graphs, to broaden the applicability of our method. Additionally, the robustness tests currently employ relatively trivial attack techniques. We plan to develop more sophisticated and realistic attack methodologies that are tailored to different domains of graph structures.

\textbf{Acknowledgement} \quad I am grateful to \href{https://github.com/Shen-Lab/LDM-3DG}{Yuning You} for his great contributions to the LDM-3DG graph generation architecture and to \href{https://github.com/lthero-big/A-watermark-for-Diffusion-Models}{lthero} for implementing Gaussian Shading, both of whom provid helpful answers to my questions during this research.

\appendix

\section{Watermark Statistical Test}
\label{app:A}

A key desideratum for a reliable watermark detector is that it provides an interpretable P-value that communicates to the user how likely it is that the observed watermark could have occurred in the data by random chance. In addition to making detection results interpretable, P-values can be used to set the threshold of detection, i.e., the watermark is ``detected'' when $p$ is below a chosen threshold $\alpha$. By doing so, one can explicitly control the false positive rate $\alpha$, making false accusations statistically unlikely.

We assume a null hypothesis $H_0$ in which the ciphertext from the latent space $c$ has an equal chance to be 0 or 1 in each bit. To test this hypothesis, we define $\eta$ as the bitwise difference between the watermarked ciphertext and the ciphertext from the molecule. The bitwise difference $\eta$ follows a binomial distribution $B(n, 0.5)$, where $n$ is the number of bits in the ciphertext.

We declare data to be watermarked if the value of $\eta$ is too small to occur by random chance. The probability of observing a value as small as $\eta$ is given by the cumulative distribution function of the binomial distribution. The P-value is thus calculated as the false positive rate (FPR) if we set the threshold as $\eta$:

\[ p = \Pr \left( B(n, 0.5) \leq \eta \right) \]

\begin{figure}[ht]
    \centering
    \includegraphics[width=0.5\textwidth]{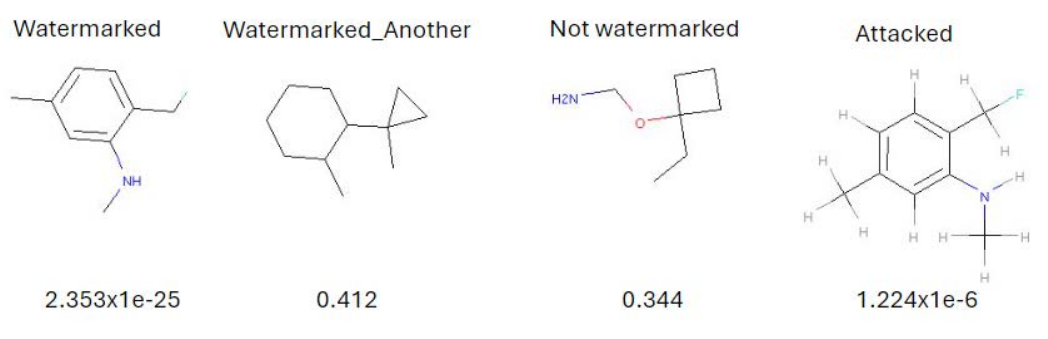}
    \caption{P-value of samples of watermarked, not watermarked, falsely watermarked, and attacked molecules}
    \label{fig:pvalue}
\end{figure}

Using this analysis, we show the calculated P-values in~\ref{fig:pvalue}  for samples drawn from various conditions:

\begin{itemize}
    \item \textbf{Watermarked}: The watermarked molecule with the detection key
    \item \textbf{Watermarked\_Another}: The watermarked molecule with another key
    \item \textbf{Not Watermarked}: Not watermarked molecule
    \item \textbf{Attacked}: Watermarked hydrogenation molecules 
\end{itemize}

\section{Evaluation Metrics}
\label{app:C}
Matrics in Table~\ref{tab:performance_comparison} \ref{tab:metrics_comparison} are calculated under the support of RDKit. RDKit is an open-source cheminformatics and machine learning software library widely used for handling and analyzing chemical data. It supports tasks such as molecular modeling, visualization, and computation of chemical properties. All the proposed functions are valid under RDkit version 2019.03.4.0.
\begin{description}
  \item[Valid] The proportion of chemically valid molecules, a molecule is considered valid if its SMILES representation smi throws no exception under. \texttt{rdkit.Chem.MolFromSmiles(smi)}
  \item[Valid\&Uni] The proportion of unique and valid
molecules among all the generated molecules, two molecules smi1 and smi2 is considered the same if \texttt{rdkit.Chem.Chem.CanonSmiles(smi1) = rdkit.Chem.Chem.CanonSmiles(smi2)}
  \item[AtomSta] The proportion of atoms in the generated
molecules that have the correct valency. An atom is considered stable If the number of its bonds matches the expected valency for that atom type. The expected valency of all possible atoms are: \texttt{{'H': 1, 'C': 4, 'N': 3, 'O': 2, 'F': 1, 'B': 3, 'Al': 3,
                 'Si': 4, 'P': [3, 5],
                 'S': 4, 'Cl': 1, 'As': 3, 'Br': 1, 'I': 1, 'Hg': [1, 2],
                 'Bi': [3, 5]}}
  \item[MolSta] The Proportion of generated molecules
in which all atoms maintain stable configurations. Atom stability is calculated the same as ``AtomSta"
\item[MW] Molecular weight, calculated via \texttt{rdkit.Chem .rdMolDescriptors.\_CalcMolWt(mol)}
\item[ALOGP] Octanol-water partition coefficient, calculated via \texttt{rdkit.Chem.Crippen.MolLogP(mol)}
\item[PSA] Polar surface area, calculated via \texttt{rdkit.Chem.MolSurf.TPSA(mol)}
\item[QED] Drug likeness, calculated via \texttt{rdkit.Chem.QED.qed(mol)}, it maps a molecule's physicochemical properties through the ADS function, weighting and summing their logarithmic values.
\item[Energy] GFN2-xTB level energy of a molecule. It's a semi-empirical quantum chemistry method used for efficiently estimating electronic structures and energies. The calculated is via \texttt{xtb.interface.Calculator} under xtb-python module.
\end{description}

\bibliographystyle{plain}
\bibliography{references}

\end{document}